\documentclass[letterpaper, 10 pt, conference]{ieeeconf}
\usepackage{amsmath,amsfonts}
\usepackage{amssymb}
\usepackage{algorithm}
\usepackage{algpseudocode}
\usepackage{array}
\usepackage[caption=false,font=normalsize,labelfont=sf,textfont=sf]{subfig}
\usepackage{textcomp}
\usepackage{stfloats}
\usepackage{url}
\usepackage{verbatim}
\usepackage{graphicx}
\usepackage{cite}
\usepackage{orcidlink}
\usepackage{xcolor}
\usepackage{bbold}
\usepackage{pgfplots}
\pgfplotsset{compat=1.18}
\usepackage{pgfplotstable}
\usepackage{float}
\usepgfplotslibrary{groupplots}
\definecolor{baseColor}{HTML}{F3F3F3}
\definecolor{baseBorder}{HTML}{D9D9D9}
\definecolor{feColor}{HTML}{FF7C80}
\definecolor{feBorder}{HTML}{FF4F53}
\definecolor{rtColor}{HTML}{84E291}
\definecolor{rtBorder}{HTML}{29B13C}

\newif\ifshownotes
\shownotesfalse   

\PassOptionsToPackage{bookmarks=false}{hyperref} 
\hypersetup{hidelinks} 

\IEEEoverridecommandlockouts
\begin{document}
\bstctlcite{MyBSTcontrol}

\title{\LARGE \bf
Tracing Back Error Sources to \\Explain and Mitigate Pose Estimation Failures
}

\author{Loris~Schneider$^{1}$, Yitian Shi$^{1,*}$, Rosa Wolf$^{1,*}$, Carolin Brenner$^{1}$, Rudolph Triebel$^{1,2}$, Rania Rayyes$^{1}$
\thanks{$^{1}$Karlsruhe Institute of Technology (KIT), Karlsruhe,
Germany.}
\thanks{$^{2}$German Aerospace Center (DLR),
Weßling, Germany.}
\thanks{$^{*}$Equal contribution.}
\thanks{This work was supported by the German Federal Ministry of Research, Technology, and Space (BMFTR) under the Robotics Institute Germany (RIG), the DFG SFB-1574-471687386 project, and the Ministry of Science, Research and Arts of the Federal State of Baden-Württemberg within the InnovationCampus Future Mobility.}
\thanks{\tt\small loris.schneider@kit.edu}%
}

\maketitle
\thispagestyle{empty}
\pagestyle{empty}

\begin{abstract}
Robust estimation of object poses in robotic manipulation is often addressed using foundational general estimators, that aim to handle diverse error sources naively within a single model. Still, they struggle due to environmental uncertainties, while requiring long inference times and heavy computation.
In contrast, we propose a modular, uncertainty-aware framework that attributes pose estimation errors to specific error sources and applies targeted mitigation strategies only when necessary. 
Instantiated with Iterative Closest Point (ICP) as a simple and lightweight pose estimator, we leverage our framework for real-world robotic grasping tasks. 
By decomposing pose estimation into failure detection, error attribution, and targeted recovery, we significantly improve the robustness of ICP and achieve competitive performance compared to foundation models, while relying on a substantially simpler and faster pose estimator.
\end{abstract}

\section{INTRODUCTION}
Object pose estimation is a central problem in robotic perception \cite{jin2025poseflow}. It refers to finding the complete six degrees of freedom (6-DoF) pose of objects in the robots environment based on sensory inputs, which is a challenging problem due to uncertainty in the environment arising from diverse sources \cite{thalhammer2024survey}. Objects might be occluded, hiding their true shape and informative geometric or color cues \cite{thalhammer2024survey}. Symmetrical and texture-less objects only provide ambiguous pose information, rendering different poses equally probable \cite{hsiao2024ambiguity}. Additionally, sensor signals might be noisy or corrupted due to perceptual failure \cite{huang2025noise}. 

Many pose estimation methods rely on depth information, but current depth sensors do not always produce high-fidelity depth images. Dark or reflective surfaces, as well as sharp edges and surrounding objects can confuse the depth sensor signal and lead to structured artifacts \cite{sajjan2020clear}.

With the progress in 6-DoF pose estimation \cite{wu2026survey}, many different methods have emerged. Among the earliest solutions are optimization-based methods, operating on point clouds. An example of a still widely used method until now is the Iterative Closest Points (ICP) \cite{besl1992icp} algorithm. It finds the transformation between two point clouds by minimizing point-to-point distances. In object pose estimation, ICP is used to find the transformation between the real pose of an object and an initial pose guess by aligning a point cloud sampled from the 3D object model to a recorded scene point cloud. However, ICP is sensitive to noise, poor initialization, and missing features of the point cloud, for example, due to occlusion \cite{dengzhi2021robust}.

Recently, deep learning based approaches have become the state of the art, showing impressive performance on different benchmarks \cite{nguyen_bop_2024}. Those methods often combine input modalities like depth with color images \cite{caraffa2024freeze}. Some use a 3D model of the object of interest as conditioning \cite{bowen_foundationpose_2024, caraffa2024freeze}, while others use reference images \cite{jung2026nemo}.
Despite the success of deep learning approaches, pose estimation remains an open problem in robotics. All current pose estimation methods expose certain failure cases due to environmental uncertainties.
While one focus of the field has been on developing powerful general pose estimation methods that are robust to all possible uncertainties, we propose a different approach: instead of a general pose estimation method, we propose developing pose estimation frameworks that can detect failure and classify error sources in the environment, and are able to deploy specific recovery strategies to mitigate catastrophic uncertainties.

\begin{figure}[t]
    \centering
\includegraphics[width=\columnwidth,trim={1.5cm 1cm 9.1cm 1.4cm}, clip]{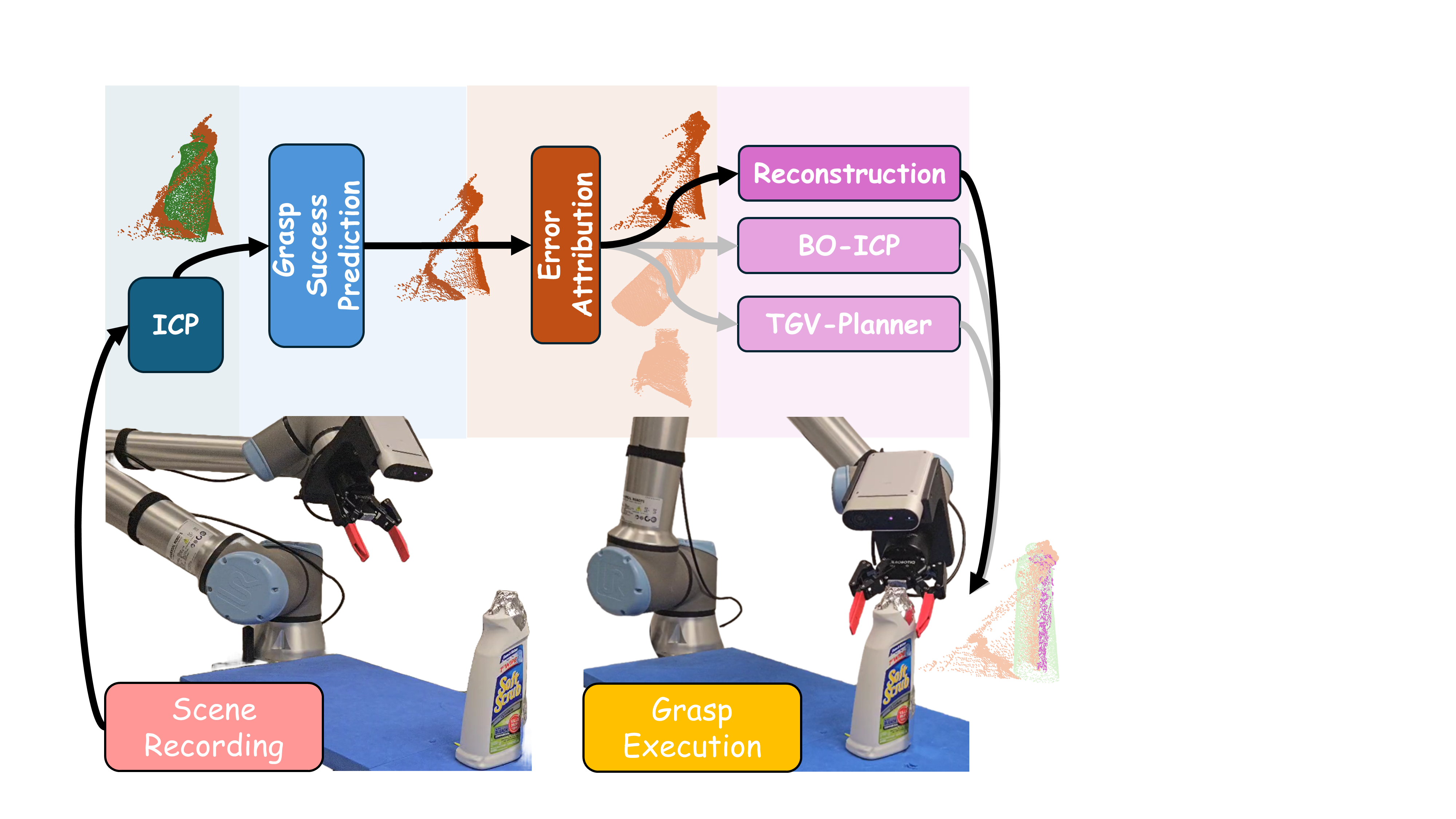}
    \caption{Overview of our framework. After the scene is recorded, ICP provides an initial pose estimate. A grasp success predictor detects possible grasp failure, which queries an error attribution system. Based on the detected error, one of the mitigation strategies will be selected. In this example, the failure is attributed to structural noise in the point cloud. A point cloud reconstruction module recovers the clean point cloud, and the pose estimate is corrected, leading to a successful grasp. Since ICP is robust to surface noise, reconstructing the rough shape is sufficient in this example.}
    \label{fig:ExampleShowcase}
\end{figure}

We design a framework, using ICP as a simple, lightweight pose estimation method and augment it with failure detection, error classification, and mitigation modules. This framework shows robustness in pose estimation for real-world robot manipulation tasks, comparable to state of the art methods. Because the underlying pose estimation method does not require a GPU, we are able to handle 
low-uncertainty cases faster and more efficiently compared to deploying FoundationPose (FP) \cite{bowen_foundationpose_2024}, a foundation model for pose estimation, while avoiding failure in high-uncertainty cases.
In summary, our contributions in this paper are:
\begin{itemize}
    \item We propose a complete grasping framework based on ICP pose estimation, grasp success prediction, error source attribution, and targeted mitigation.
    \item We propose a novel point cloud reconstruction method, specifically designed to recover ICP pose estimation errors due to real-world noise.
    \item We extend previous work on ICP error source attribution by applying a transformer-based classifier, significantly improving attribution accuracy in real-world settings, and solving the confusion problem between occlusion and bad initialization.
    \item We evaluate our framework on real-world grasping tasks, achieving similar and, in the case of noise, higher grasp success rates compared to using FoundationPose \cite{bowen_foundationpose_2024} as pose estimator.
\end{itemize}

\section{RELATED WORK}
\label{sec:related_work}
Our framework builds on and integrates ideas from three research directions: object pose estimation, error identification and attribution, and mitigation strategies for pose estimation failures. While these have been studied individually in prior work, we combine them into a unified framework.

\subsection{Object Pose Estimation}
Object pose estimation aims to find 6-DoF poses for objects. An established method that solves this problem on the basis of point clouds is ICP \cite{besl1992icp}. It tries to find the rigid transformation between two point clouds by aligning the source point cloud to the target point cloud. In robotic object manipulation, the source point cloud is usually sampled from a 3D object model, and the target point cloud is the object of interest segmented from the recorded scene point cloud.

Despite its efficiency, classical ICP is sensitive to poor initialization, missing features due to occlusion, and sensor noise, and may converge to local minima under such conditions \cite{besl1992icp, kolpakov2023einit}. These limitations restrict its robustness in unstructured real-world environments.


A highly regarded example of a deep learning based method is FoundationPose (FP) \cite{bowen_foundationpose_2024}, a general foundation model for object pose estimation. FP uses image encoders to first detect the object in the input RGB-D image and then samples pose hypotheses around its rough location. The pose hypotheses are refined by rendering the object model as it would appear from the camera under each pose hypothesis, generating synthetic RGB-D images. Those are then aligned to the original input image by refining the pose hypotheses. Finally, a scoring network uses multi-head self-attention across all pose hypotheses to generate a score for each pose hypothesis. The pose with the highest score is then selected.

Although demonstrating strong performance in diverse scenarios, the accuracy of the pose estimate can deteriorate severely in cases of heavy occlusion, ambiguous object symmetries and sensor noise \cite{bowen_foundationpose_2024, nguyen_bop_2024}. Moreover, FP relies on computationally expensive hypothesis rendering and neural inference, requiring significantly higher runtime compared to classical methods \cite{nguyen_bop_2024}.

\subsection{Error Identification and Attribution}
A major step towards mitigating the sources of uncertainty in ICP is classifying them. Three sources of uncertainty in ICP registration are \textit{sensor noise}, \textit{bad pose initialization}, and \textit{occlusion} \cite{censi2007icpcov}. In previous work \cite{qin2024towards}, SHAP \cite{lundberg2017shap} is used to attribute and quantify different sources of uncertainty on a set of registered point clouds.  GP-CA \cite{gaus2025human} encodes the registered point cloud with a DGCNN and combines uncertainty attribution and quantification, with an active learning approach, to continuously adapt the uncertainty classifier. However, both frameworks only add Gaussian noise to simulate sensor noise, which is not a realistic simulation of most modern depth sensors \cite{pomerleau2012noise}. Moreover, neither method implements mitigation strategies. In contrast, we implement and evaluate error mitigation on a real-robot setup, also predicting grasp failure before grasp execution.

\subsection{Mitigation}
Error sources like noise, occlusion, and bad initialization are not exclusive to our setting and have been addressed individually in prior work. In the following, we outline relevant approaches to mitigate them.

\paragraph{Point cloud reconstruction and denoising}
As one of the mitigation strategies, point cloud denoising aims to remove measurement noise and outliers while preserving geometric details. Learning-based denoising has benefited from the recent success of point-based neural networks \cite{qi_pointnet_2016, wang_dgcnn_2019}. For instance, PointCleanNet predicts point-wise corrections after rejecting outliers \cite{rakotosaona2019pointcleannet}, while the score-based method SnowFlake \cite{luo2021scoredenoise} iteratively moves noisy points toward the surface using estimated gradients. In addition, point-based reconstruction methods have further improved the structural fidelity of denoising by introducing grid-based intermediates \cite{xie2020grnet} or transformer-based set-to-set translation models, such as PoinTr \cite{yu2021pointr, yu2023adapointr}. 

However, most approaches are evaluated under simplified synthetic perturbations, such as Gaussian noise or randomly injected outliers. These do not faithfully reflect the structured and incomplete corruption produced by real depth sensors for robotic applications.

\paragraph{ICP initialization}
Since ICP is based on optimization, it is susceptible to converging to suboptimal local minima. This behavior is sensitive to the initialization of the algorithm, which is why methods have been developed to find an initialization that avoids suboptimal convergence of ICP. E-Init \cite{kolpakov2023einit} finds an initial alignment based on calculating the principal axes of the source and the target point cloud.

Go-ICP \cite{yang2016goicp} implements a Branch-and-Bound search in 3D-space for global optimal alignment and uses ICP to evaluate transformation candidates. Similarly, BO-ICP \cite{biggie_bo_icp_2023} uses a Bayesian optimization of the initial transformation.

\paragraph{Occlusion-aware view planning for manipulation}
For robotic manipulation in highly cluttered scenes, view selection is crucial for reducing occlusions. Prior work formulates this as a next-best-view (NBV) problem \cite{zeng2020view}, seeking viewpoints that maximize target visibility for downstream perception and manipulation. Existing approaches either optimize information gain (IG) for discrete view selection \cite{marques2025map, breyer2022closed} or perform continuous view planning with spatial-semantic-aware objectives. In particular, Viso-Grasp~\cite{shi_visio_grasp_2025} models explicit object-centric spatial relationships through semantics and guides viewpoint optimization with a pre-constructed velocity field. This design enables continuous observation along the view trajectory and facilitates multi-view information integration tailored for uncertainty-aware grasping approaches \cite{shi2025vmf, shi2024uncertainty}.

\begin{figure*}[t]
    \centering
    \includegraphics[width=\textwidth, trim={1cm 8cm 0.5cm 0cm},clip]{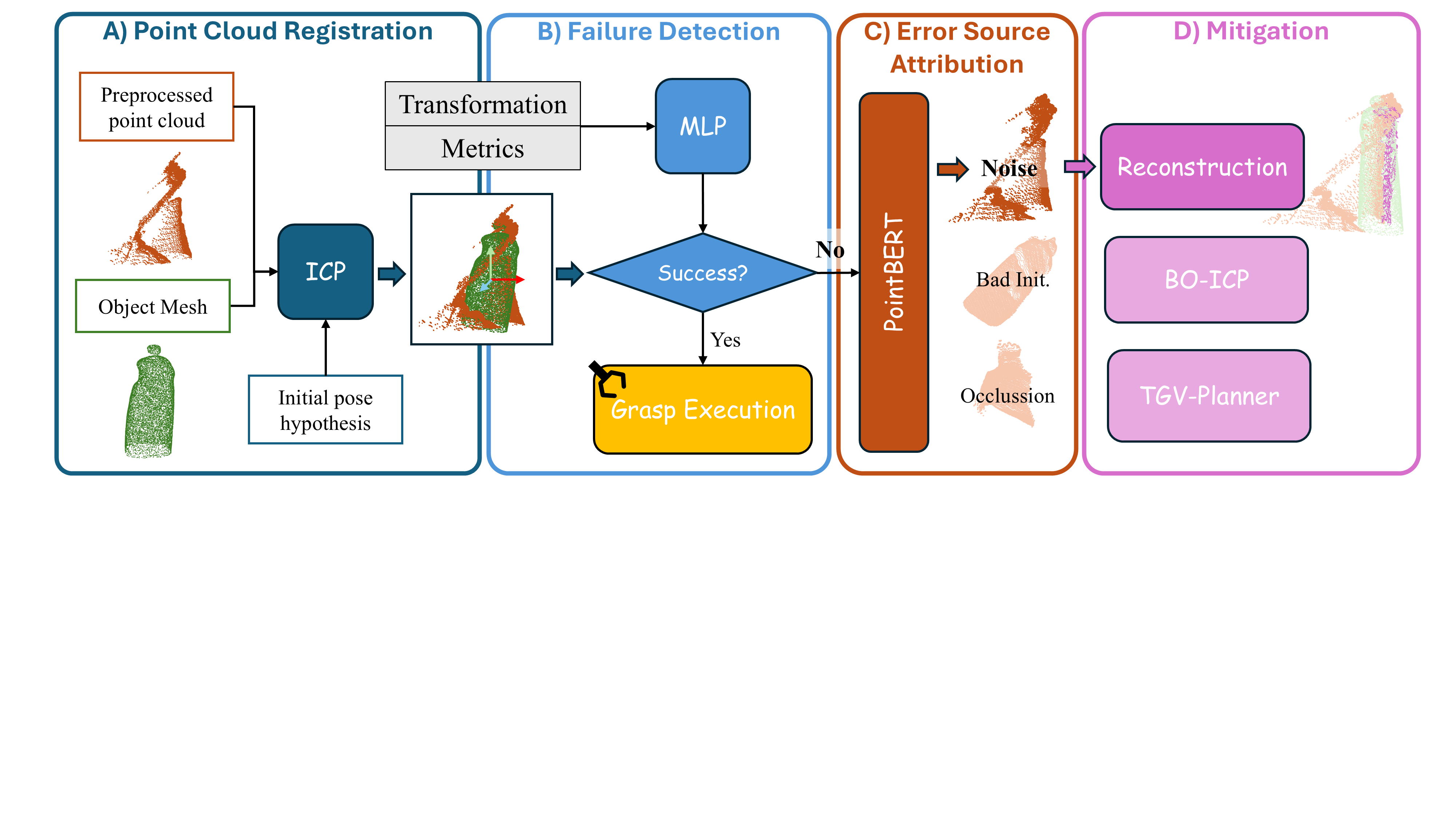}
    \caption{Overview of the deployed framework. (A) A point cloud sampled from the object mesh is aligned to the recorded scene point cloud via ICP. (B) An MLP predicts grasp success based on the calculated transformation estimate and alignment metrics. (C) In the case of predicted grasp failure, a PointBERT \cite{yu2021pointbert} classifier attributes the failure to a specific source, here shown for noise. (D) Based on the classification, targeted mitigation strategies - BO-ICP \cite{biggie_bo_icp_2023}, the TGV-Planner \cite{shi_visio_grasp_2025}, and a custom point cloud reconstruction model - are applied to recover the pose estimate. In the illustrated case, the clean point cloud is reconstructed from the noisy point cloud and used for ICP alignment, which corrects the pose estimate.}
    \label{fig:SystemOverview}
\end{figure*}

\section{METHODOLOGY}
\label{sec:method}
We leverage our framework for pose estimation in robotic manipulation tasks. We apply a simple and lightweight point-to-point ICP for the pose estimation, predict grasp success, and deploy targeted mitigation strategies only when specific error sources are present. Figure~\ref{fig:SystemOverview} shows the overall system overview, which represents a framework for pose estimation and uncertainty-aware manipulation using 3D point cloud data.

\subsection{Scene Recording and Prep-processing}
Initially, the environment is recorded using a depth sensor to generate a raw point cloud. The recorded scene is then cropped, and the segmented area of the regarded point cloud is focused to reduce computational complexity and remove irrelevant background structures. The result is a preprocessed recorded point cloud $P\in \mathcal{R}^{N\times3}$ with $N$ points. In the next step, a point cloud $P_M$ is sampled from the 3D mesh model of the target object and aligned to the recorded point cloud using the ICP algorithm and an initial pose hypothesis. Then, the result of the ICP is evaluated and the failure detection and uncertainty attribution follows.

\subsection{Failure Detection and Error Source Attribution}
\label{sec:error_attribution}
After ICP registration, we apply a lightweight classifier to predict whether the registration result is likely to yield a successful grasp or not. To this end, we calculate multiple quality metrics based on the aligned point clouds.

In the following, we use~$||\cdot||$ to denote Euclidean distance.
The fitness $F_\tau$ refers to the ratio of mesh points that lie within a distance threshold $\tau$ to a scene point:
\begin{equation*}
    F_\tau = \frac{\displaystyle\sum_{p_m \in P_M} \mathbf{1}_{\{\exists p \in P: ||p_m - p|| < \tau\}}}{|P_M|}.
\end{equation*}
We compute two fitness ratios, with a distance threshold of one centimeter and two centimeters, respectively.
The root mean squared error $\mathrm{RMSE}_{\mathrm{inlier}}$ is calculated as the average distance over all mesh points~$p_m \in P_M$ to their nearest neighbors~$\mathrm{NN}_{p_m}$ in~$P$, within a third threshold~$\tau$: 
\begin{equation*}
    \mathrm{RMSE}_{\mathrm{inlier}, \tau} = \sqrt{\frac{\displaystyle \sum_{p_m \in P_M} \mathbf{1}_{\{||p_m - \mathrm{NN}_{p_m}|| < \tau\}} ||p_m-\mathrm{NN}_{p_m}||^2}{\displaystyle\sum_{p_m \in P_M} \mathbf{1}_{\{||p_m - \mathrm{NN}_{p_m}|| < \tau\}}}}.
\end{equation*}

Finally, we calculate a distance measure between the aligned point clouds as:
\begin{align*}
    D_{P_M\rightarrow P} &= \frac{1}{|P_M|}\sum_{p_m \in P_M}\min_{p \in P}||p_m - p||,\\
    D_{P\rightarrow P_M} &= \frac{1}{|P|}\sum_{p \in P}\min_{p_m \in P_M}||p - p_m||.
\end{align*}
Our failure predictor is implemented as a multi-layer perceptron (MLP) that takes fitness, inlier RMSE, point cloud distances, as well as the transformation estimate $T$ as input and predicts a binary success label $S$:
\begin{equation*}
    S = \mathrm{MLP}(F_{\tau_1}, F_{\tau_2}, \mathrm{RMSE}_{\mathrm{inlier}, \tau_3},D_{P_M\rightarrow P}, D_{P\rightarrow P_M},T).
\end{equation*}
If an alignment is predicted to be unsuccessful, we don't execute a grasp but proceed to the uncertainty attribution which is implemented as a PointBERT \cite{yu2021pointbert} classifier that takes the recorded scene point cloud $P$ as input and outputs class probabilities over the discrete uncertainty classes.

\begin{figure*}
    \centering
    \includegraphics[width=\textwidth, trim={0.5cm 4.9cm 1.6cm 4cm}, clip]{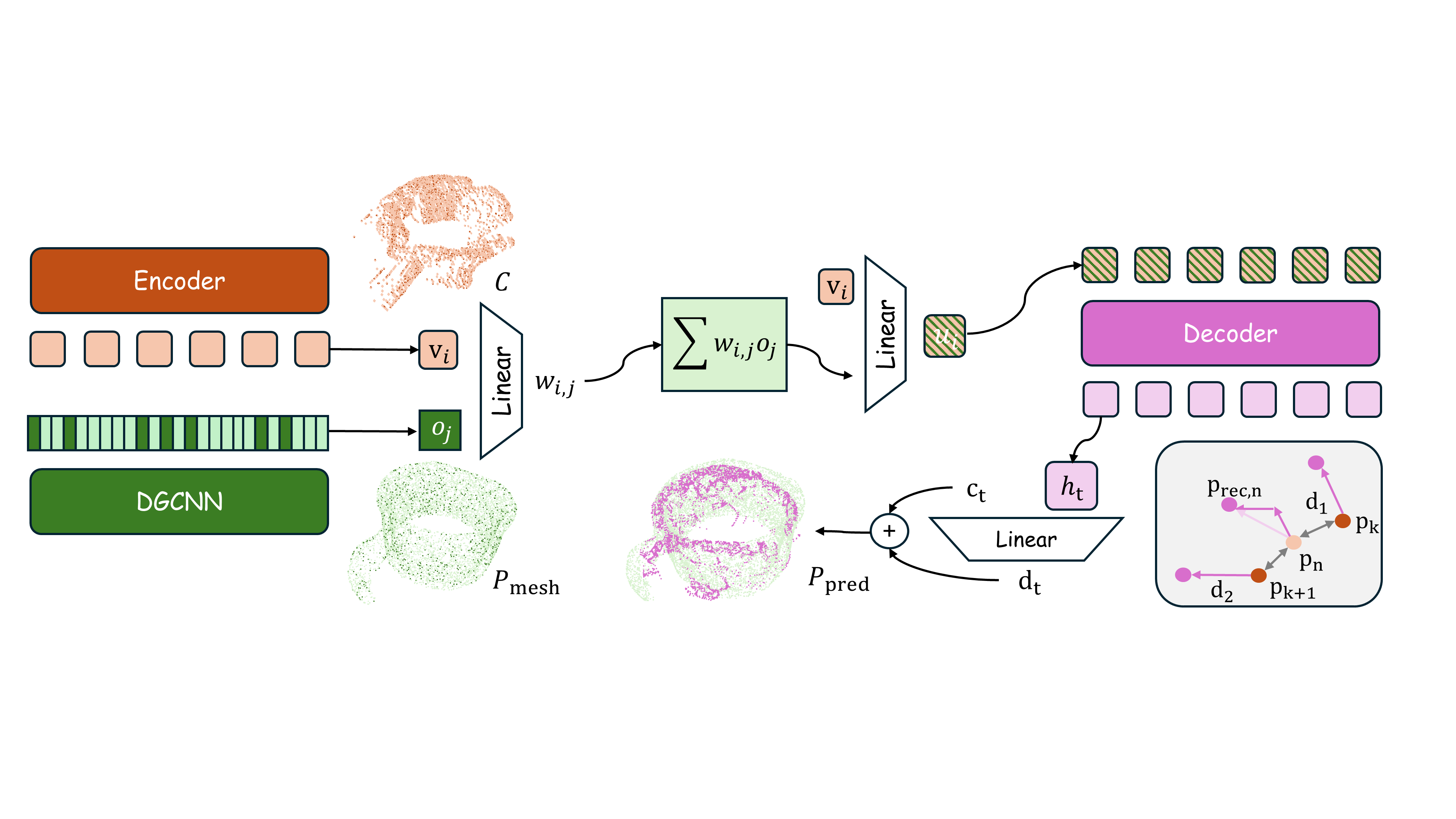}
    \caption{Point cloud reconstruction module. A transformer based encoder creates tokens from point patches of the noisy point cloud, while a DGCNN \cite{wang_dgcnn_2019} encodes point-wise features of the mesh. From the point-wise mesh features, a subset is selected via FPS. The tokens from the noisy point cloud are merged with the selected point-wise mesh features by calculating a weighted sum with learned weights. An encoder predicts displacement vectors for each patch center point of the noisy point cloud, which is propagated to the surrounding points.}
    \label{fig:reconstruction}
\end{figure*}

\subsection{Mitigation Strategies}
\label{sec:mitigation}
Once the uncertainty is attributed to a specific class, we deploy targeted mitigation strategies in order to enable ICP to find a better solution.

\subsubsection{Point Cloud Reconstruction}
Since ICP optimizes mean correspondence distances, it is robust to Gaussian jitter of points. However, noise artifacts, consisting of large structural displacements, can deteriorate the ICP result. Under such noise, on the other hand, it is not trivial to recover the original point cloud. To tackle this, we apply a custom model based on the PoinTr \cite{yu2023adapointr} architecture to solve this reconstruction task as illustrated in Fig. \ref{fig:reconstruction}. 

Specifically, starting from a distorted scene point cloud $P$ with $N$ points, we aim to recover a reconstructed point cloud $P_{\mathrm{rec}}\in \mathcal{R}^{N\times3}$. Similar to \cite{yu2023adapointr, yu2021pointbert}, we identify $I$ center points of the scene point cloud using Farthest Point Sampling (FPS): 
\[
C = \text{FPS}(P), \quad C\in \mathcal{R}^{I\times3}.
\]

After that, k-nearest-neighbor (kNN) is applied to categorize points into $I$ clusters. The point cloud clusters are further tokenized by a lightweight PointNet-style \cite{qi_pointnet_2016} encoder and an MLP as positional encoder into $I$ tokens: 
\[
\mathcal{F} =\text{MLP}\circ\text{PointNet}\circ\text{kNN}(P,C), \quad \mathcal{V} = \text{Tr}_\text{enc}(\mathcal{F}).
\] 
Analogous to \cite{yu2021pointbert}, the scene point cloud tokens are processed by a transformer-based encoder $\text{Tr}_\text{enc}$.

In parallel, we encode point-wise features from the mesh point cloud $P_M$ with a Dynamic Graph Convolutional Neural Networks (DGCNN) \cite{wang_dgcnn_2019}. We select a subset of $J$ points from $P_M$ via FPS and collect their point-wise features $\mathcal{F}_M$ as proxies $\mathcal{O}$ for the mesh point cloud:
\begin{align*}
    \mathcal{O} &= \text{DGCNN} \circ\mathrm{FPS}(P_M) 
\end{align*}
Using the scene encoder output tokens $\mathcal{V}$ and the mesh proxies $\mathcal{O}$, we compute a single weight $w_{i,j}$ for each token and proxy pair:
\begin{equation*}
    w_{i,j} = \mathrm{Linear}({v}_i, {o}_j), \ \text{where} \ {v}_i \in \mathcal{V}, {o}_j \in \mathcal{O},
\end{equation*}
where $\mathrm{Linear}$ refers to a linear layer. Using the pair-wise weights, we fuse each token with the proxies by calculating a weighted sum:
\begin{equation*}
    {u}_i = \mathrm{Linear}({v}_i,\sum_{j\in1,...,J}w_{i,j}{o}_j)
\end{equation*}

The resulting tokens $\mathcal{U} = \{{u}_i\}_{i=1,...,I}$ are fed into a transformer-based decoder $\text{Tr}_\text{dec}$ which outputs the final tokens $\mathcal{H} = \text{Tr}_\text{dec}(\mathcal{U})$. The reconstruction $  p_{\mathrm{rec}, c} \in P_\text{rec}$ of a center point $ p_c \in C$ is achieved by predicting a displacement vector $ d_c$:
\begin{align*}
     d_c &= \mathrm{Linear}( p_c, {h}_c)  \in \mathcal{R}^3, \ \text{where} \  h_c \in \mathcal{H},\\
     p_{\mathrm{rec}, c} &=  p_c +  d_c.
\end{align*}
Due to the robustness of ICP to small surface jitter, it is sufficient to reconstruct the rough point cloud shape. Therefore we refrain from predicting fine-grained point-wise reconstruction like in \cite{yu2023adapointr} and instead calculate the final reconstruction of each remaining point $p_{\mathrm{rec},n} \in P \setminus P_C$ by applying the distance-weighted average displacement of its $K$ nearest center points:
\begin{equation*}
     p_{\mathrm{rec},n} =  p_n + \sum_{k\in K}\frac{\alpha_{n,k}}{\sum_{k' \in K} \alpha_{n, k'}}d_k,  \ \text{where} \ \alpha_{n,k} = ||p_n - p_k||^{-1}.
\end{equation*}
The final transformation estimate can then easily be calculated by using ICP to align the mesh point cloud to the reconstructed point cloud.

\subsubsection{Registration Initialization}
Even if the scene point cloud is of high quality, ICP commonly fails when converging to a suboptimal local minimum. Whether this happens is highly dependent on the initial transformation estimate \cite{yang2016goicp}. In our case, we expect the object of interest to have a certain pose and therefore use an identity transformation as the initial estimate. If the real object pose differs substantially from our initial hypothesis, ICP is likely to converge to a suboptimal local minimum.

To mitigate this, we use BO-ICP \cite{biggie_bo_icp_2023}, a Bayesian optimization on the initial transformation estimate $T = (x, y, z, \psi, \theta,\varphi)$ where $x,y,z$ are the translation components and $\psi, \theta, \varphi$ are roll, pitch and yaw angles. A transformation estimate is evaluated by running ICP, resulting in the transformed mesh point cloud~$P_{M, T} = \mathrm{ICP}(P_M, P, T)$. As an objective function~$\mathcal{L}$, we use a measure of overlap between the aligned point clouds:
\begin{align*}
    \mathcal{L} = \max_{T} & \begin{cases}
            \Omega(T), & F(P_{M,T}, P) >0\\
            \Omega_{D^-}(T), & F(P_{M,T},  P) = 0
        \end{cases}, \\
    \Omega_{D^-} &= -||\frac{1}{|P_{M,T}|}\sum_{p_m \in P_{M,T}}p_m - \frac{1}{|P|}\sum_{p\in P} p||\\
    \Omega &= F - \mathrm{RMSE}_{\mathrm{inlier}}.
\end{align*}

If there is no overlap between the point clouds, we use the negative distance~$\Omega_{D^-}$ between the point cloud center points. Otherwise, using the fitness~$F$ and~$\mathrm{RMSE}_\mathrm{inlier}$, the measure~$\Omega$ counts the ratio of points in~$P_{M, T}$ that are aligned with~$P$, while also penalizing higher distance from point cloud~$P$. 

By including the distance between the point clouds in the objective function, BO-ICP is able to recuperate large translation errors because it searches the workspace for initial transformations that result in maximal overlap between the point clouds. It is also able to find favorable rotations such that the alignment of both point clouds can be maximized, once overlap is found \cite{biggie_bo_icp_2023}. Pose errors that are not reflected in the objective function can not be recovered by BO-ICP. In our case, those include, for example, ambiguities due to symmetries, missing parts in the point cloud, or strong noise artifacts.

\subsubsection{Occlusion}
Occluding objects, blocking the view of the depth sensor, result in partial point clouds with missing geometries. This can negatively affect ICP performance. In this case, we move the end-effector camera to a different viewpoint in order to get a clear view of the object of interest. 

Specifically, we adopt the dynamic next-best-viewpoint planner introduced in \cite{shi_visio_grasp_2025}, termed the Target-Guided View Planner (\emph{TGV-Planner}). This view planner constructs a velocity field over a pre-defined hemisphere above the workspace, providing guidance for occlusion avoidance. Compared to the discrete next-best-view selection strategy as \cite{breyer2022closed}, the velocity field-based approach formulates the view optimization as a time-dependent trajectory planning problem, which better suits the integration with the velocity control of the camera viewpoint.

\subsection{Training details}
\label{sec:method_training}

\begin{figure}[t]
    \centering
\includegraphics[width=0.9\columnwidth, trim={2.1cm 3.2cm 2.6cm 6.4cm}, clip]{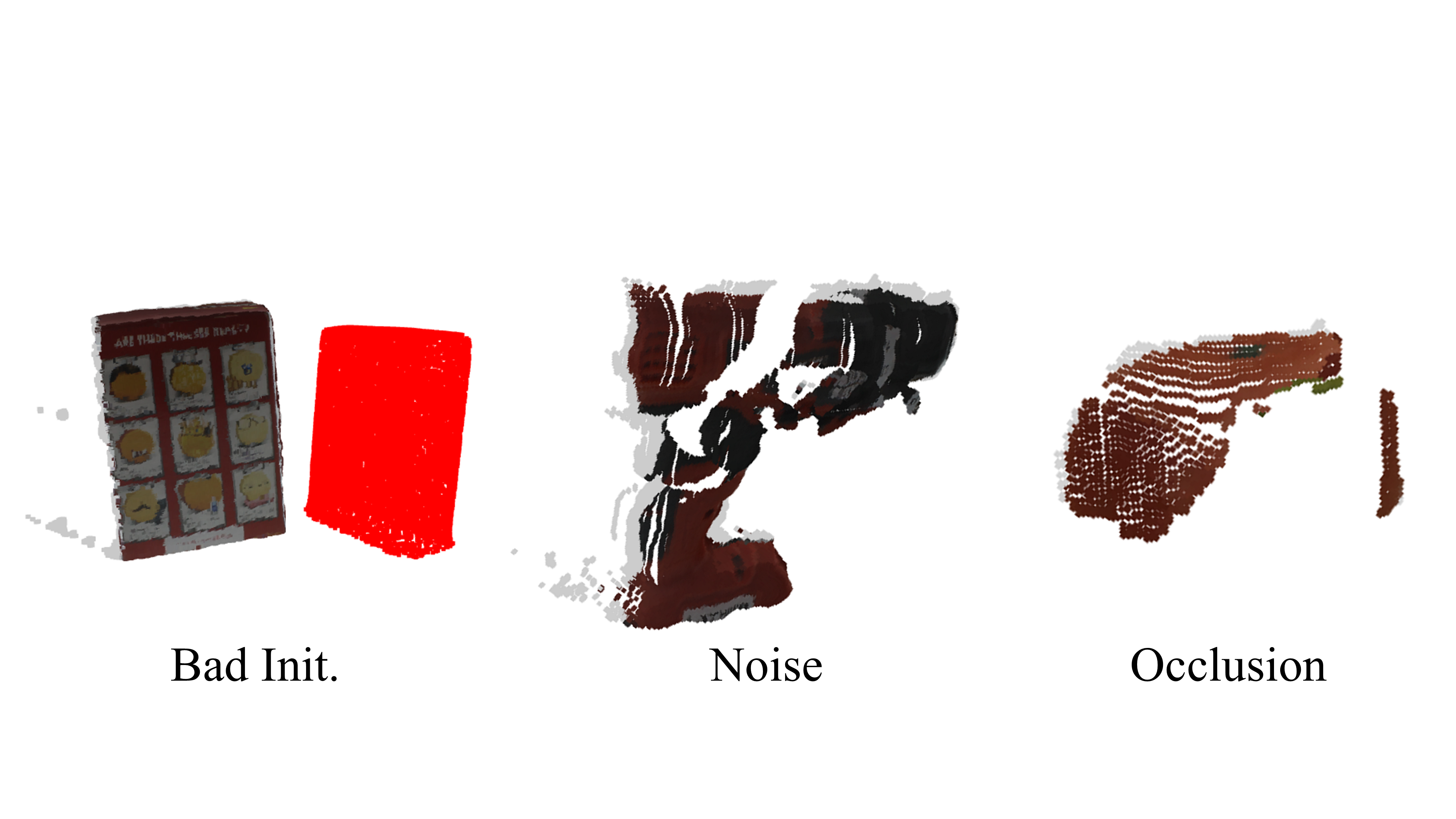}
    \caption{Synthetic scenes showing the error cases created using SAPIEN \cite{Xiang_2020_SAPIEN}. For bad initialization, the initial pose hypothesis is shown in red.}
    \label{fig:synthetic_data}
\end{figure}

For training the PointBERT-based \cite{yu2021pointbert} error source attribution described in \ref{sec:error_attribution}, a combination of synthetic data and real-world data is used. The synthetic data is generated in SAPIEN \cite{Xiang_2020_SAPIEN}, as it offers realistic simulation of a stereo depth sensor, reducing the sim-to-real gap.
For training the classifier, we created a synthetic dataset of 4.700 scenes using the YCB \cite{calli_ycb_2015} objects. In each scene, we replicate a specific error case. Figure~\ref{fig:synthetic_data} shows examples of the created synthetic scenes for each error case. Additionally, we collect 200 samples from 81 real-world scenes showing the error cases, from which we separate 150 samples for training. A detailed description of our real-world data collection is provided in Section \ref{sec:experiment_setup}. We train on both synthetic and real-world samples, gradually increasing the ratio of real-world samples as training progresses. This curriculum allows us to achieve high accuracy on real-world data while minimizing the burden of real-world data collection.

For training the point cloud reconstruction described in \ref{sec:mitigation}, we create 55.000 synthetic scenes in a similar fashion. For each scene, we record a noisy point cloud with SAPIEN's implementation of a realistic depth sensor and a ground truth point cloud using ray-casting. The noisy point cloud is additionally distorted by applying random translations to randomly shaped patches. Again, we gradually inject 300 samples of real-world distorted point clouds during training. As loss function, we use the Chamfer distance between the distorted point cloud $P$ and the ground truth point cloud $P_{\mathrm{gt}}$ 
\begin{equation*}
    D_{\mathrm{Chamfer}} = D_{P_{\mathrm{gt}\rightarrow P}} + D_{P\rightarrow P_{\mathrm{gt}}}.
\end{equation*}

\section{EXPERIMENTAL RESULTS}
For evaluation of our framework, we provide ablations for its individual components, as well as experiments on real-world grasping tasks. 

\begin{figure*}[t]
    \includegraphics[width=\linewidth, trim={0.2cm 6.6cm 0.2cm 0.4cm}, clip]{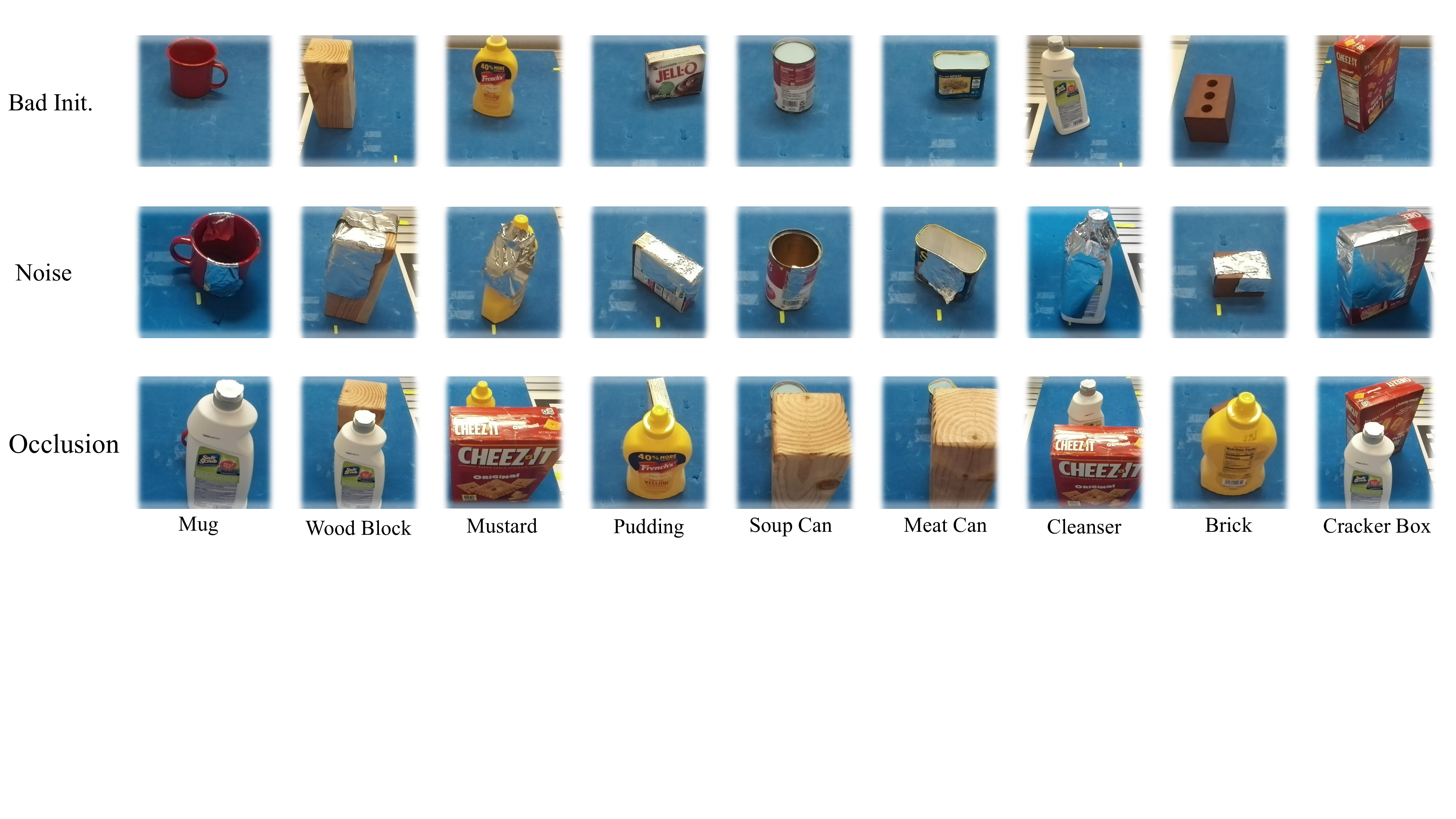}
    \caption{Overview of the real-world scenes used for evaluation and training. For each object, we show one example of each error case.}
    \label{fig:experiments_overview}
\end{figure*}

\subsection{Experiment Setup}
\label{sec:experiment_setup}
To evaluate our framework and its components on real-world cases, we create scenes using household objects, replicating the error cases. Specifically, we used nine objects from the YCB \cite{calli_ycb_2015} dataset and set a fixed approach and grasp poses, as well as initial pose hypotheses for each object. Figure \ref{fig:experiments_overview} shows examples for each object and each error case.

\textbf{Bad Initialization.}
To create ICP failure due to bad initialization, we place the object far from its initial pose hypothesis, forcing ICP to realize large corrections.

\textbf{Noise artifacts.}
 In order be able to compare point clouds of the same object with realistic structural noise to point clouds without it, we wrap parts of the object surface in aluminum foil as shown in Figure \ref{fig:experiments_overview}. This creates removable reflective surfaces that deteriorate depth estimation while preserving the object shape without affecting the underlying object surface. The Tomato Soup Can and the Potted Meat Can provide the additional option to record them either in an opened or a closed state. When opening the can, the hollow interior and the metallic surface inside the can result in particularly challenging but realistic depth recordings.

\textbf{Occlusion.} Occlusion is easily created by placing an object of larger or similar size between each object and the camera viewpoint. As observable in Figure \ref{fig:experiments_overview}, we make sure that a fraction of the object remains visible at all times.

We create an initial dataset of 81 scenes, containing each object three times in each uncertainty case and record multiple viewpoints per scene. From this, we reject viewpoints that don't provide satisfactory views on the object, as well as point clouds with less than 2048 points, to arrive at a balanced dataset of 200 samples. From this, we separate 50 samples for evaluation containing different object than the training set.

To evaluate the grasp success prediction, we additionally perform a pose estimation and grasp execution for each scene and create an additional three scenes per object in which we apply no uncertainty case and place the object close to its initial pose hypothesis provoking successful grasps. 

\begin{figure}[t]
    \centering\includegraphics[width=0.5\linewidth, trim={7.7cm 1.9cm 15.5cm 6cm}, clip]{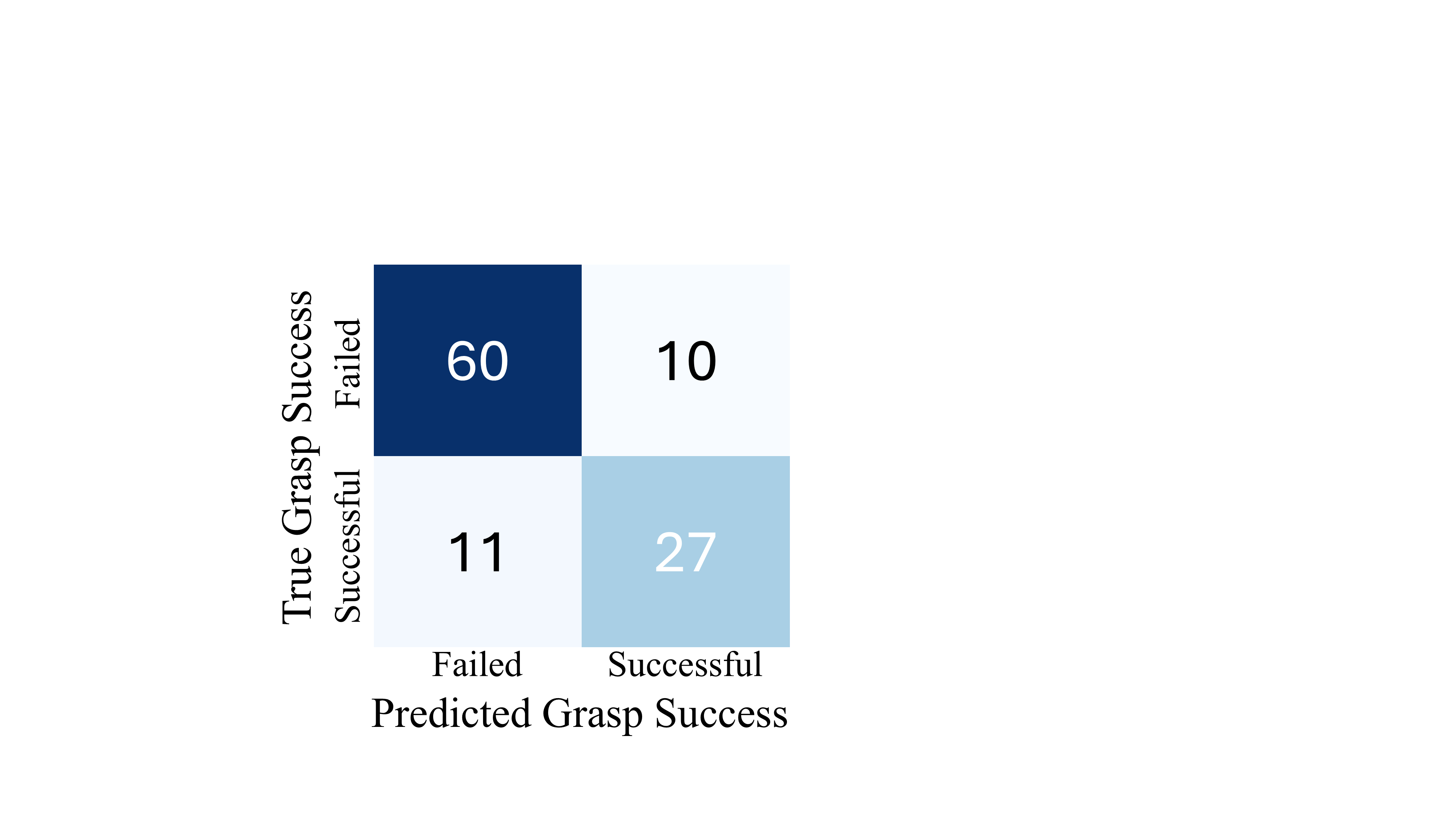}
    \caption{Confusion matrix of the failure detection MLP evaluated on 108 grasping tasks.}
    \label{fig:error_predictor_cm}
\end{figure}

\subsection{Failure Detection}
Initially, we evaluate our failure detection MLP on 10.000 scenes created using the SAPIEN \cite{Xiang_2020_SAPIEN} simulation environment, as described in \ref{sec:method_training}. We run ICP on those scenes and report the error between estimated and ground truth pose using the Average Distance of Model Points (ADD) \cite{hinterstoisser2012add} \[
\mathrm{ADD}(R,t) = \frac{1}{|P_M|}\sum_{p_m \in P_M}||(Rp_m + t) - (\hat{R}p_m + \hat{t})||
\] 
where $(R,t)$ describes the rotation and translation of the ground truth pose and $(\hat{R}, \hat{t})$ the rotation and translation of the estimated pose.

Since we do not execute grasps in simulation, we evaluate our failure detection MLP on predicting whether the ADD error lies above or below a threshold. After training on 7.000 scenes and evaluating on 3.000 scenes, we achieve an accuracy of 92\% on synthetic data.

Finally, we record the grasp success of 252 diverse real-world grasping tasks using ICP pose estimation and train our failure detection MLP on the results. We evaluate on 108 grasping tasks, containing our specific uncertainty cases, and achieve a grasp success prediction accuracy of 80.5\% and report the confusion matrix depicted in Figure \ref{fig:error_predictor_cm}.

\subsection{Uncertainty Attribution}
We train and evaluate PointBERT for error attribution on the same dataset used in \cite{gaus2025human} and report the results in Tab.~\ref{tab:confusion-rate}. We observe that PointBERT not only yields higher accuracy, but also almost completely mitigates the confusion between the bad initialization class and the occlusion class that was previously reported in \cite{gaus2025human}.

\begin{table}[t]
    \centering
     \caption{Accuracy of uncertainty attribution and confusion rates of bad initialization (Bad Init.) and occlusion compared to the results in \cite{gaus2025human}.}
    \begin{tabular}{lcccc}
    \hline\\[-1.5ex]
          \textbf{Model} & \textbf{Mean Acc.} & \shortstack{\textbf{True Bad Init.}\\ \textbf{Pred. Occlusion}} &  \shortstack{\textbf{True Occlusion}\\\textbf{Pred. Bad Init.}} \\
          \hline\\[-1.5ex]
         \textbf{PointBERT} &  99.47\% & 1.02\% & 0.61\%\\
         DGCNN \cite{gaus2025human} & 85.20\% & 20.60\% & 23.00\%\\ 
         \hline
    \end{tabular}
    \label{tab:confusion-rate}
\end{table}
Additionally, we evaluate PointBERT on 50 unseen samples of real-world scenes showing the uncertainty cases as described in \ref{sec:experiment_setup}. We compare the accuracy of PointBERT to using a DGCNN and a PointNet \cite{qi_pointnet_2016} in five training and evaluation rounds in Tab.~\ref{tab:accuracy}. On real-world cases, PointBERT achieves a mean attribution accuracy of 83.83\%. This significantly outperforms DGCNN and PointNet, which achieve 71.49\% and 50.64\% mean attribution accuracy, respectively. This validates our choice of applying a transformer-based model like PointBERT over point encoder models like DGCNN and PointNet.

\begin{table}[t]
    \centering
     \caption{Accuracy of error attribution on real-world scenes.}
    \begin{tabular}{lcc}
    \hline\\[-1.5ex]
          \textbf{Model} & \textbf{Mean Acc.} & \textbf{Std. Dev.}\\
          \hline\\[-1.5ex]
         \textbf{PointBERT} &  83.83\% & 2.27\%\\
         DGCNN & 71.49\% & 2.55\% \\ 
         PointNet & 50.64\% & 3.40\%\\
         \hline
    \end{tabular}
    \label{tab:accuracy}
\end{table}

\subsection{Framework Evaluation and Baseline Comparison}
Finally, we evaluate our complete framework on grasping tasks containing each error case. For each error case, we create 20 real-world scenes and execute a grasping task using our framework of ICP pose estimation, success prediction, error attribution and mitigation, and grasp execution. As a baseline, we compare grasp success rates to using FP \cite{bowen_foundationpose_2024} for pose estimation instead of our framework.

\begin{table}[!t]
    \centering
    \caption{Grasp success comparison of our framework against FP \cite{bowen_foundationpose_2024} for each error case.}
    \label{tab:gras_p_transitions}
        \begin{tabular}{lccc}
        \hline\\[-1.5ex]
            \textbf{Error Case} & \textbf{ICP success} & \shortstack{\textbf{ICP\&Mitigation}\\ \textbf{success}} & \textbf{FP success}\\
            \hline\\[-1.5ex]
            Bad Init. & 0\% & 55\% & \textbf{85\%}\\
            Noise & 15\% & \textbf{80\%} & 30\%\\
            Occlusion & 10\% & \textbf{70}\% & \textbf{70\%}\\
            \hline
        \end{tabular}
\end{table}

\paragraph{Grasp success}
The results are displayed in Table \ref{tab:gras_p_transitions}. Generally, the targeted mitigation of uncertainty cases led to a significant increase in grasp success, while still only relying on ICP for pose estimation. As FP does not rely on an initial pose hypothesis, it is not affected by bad initialization, achieving 85\% success rate in this case compared to 55\% success rate using our framework. The success rate dropped for occlusion cases but was still as high as 70\%, on par with our framework. In the case of noise, FP was not able to predict a sufficiently accurate pose for grasp success in most cases, with a grasp success rate of only 30\% compared to 80\% success rate of our framework. Overall, the grasp success rate of FP was 61.7\% and as such only slightly higher than the 60\% achieved by our framework.

We report the success rate or ICP without mitigation separately in Table \ref{tab:gras_p_transitions}. The sensitivity of ICP to the investigated error cases is evident with 0\% grasp success for bad initialization, 10\% for occlusion, and 15\% for noise.




The results show that our approach of detecting uncertainty, attributing it to specific sources, and performing targeted mitigation can significantly increase the performance of the otherwise limited ICP. Our framework was able to achieve comparable grasp success rates to FP, all while relying only on point clouds as data modality and ICP as a lightweight and simple classical pose estimator.

\paragraph{Computation time and energy consumption}
The difference in complexity between ICP and FP is further evident in the computation time and energy consumption both need to provide a pose estimate. To illustrate this, we ran pose estimation using ICP and FP on 10.000 samples from the YCB-Video \cite{xiang2018posecnn} dataset. FP ran on a NVIDIA GeForce RTX 5080 GPU, while ICP was executed on an Intel Core i7-14700K (Raptor Lake Refresh, 8P+12E cores, 28 threads). Over all samples, FP had a mean inference time of 1.36 seconds and a mean inference energy consumption of 377.81 Joule per inference. ICP had an execution time of only 0.06 seconds while consuming 9.85 Joules on average, making it roughly 22.7 times faster while using 38.4 times less energy than FP.
This indicates that our framework is able to provide much faster and efficient pose estimates for cases where no mitigation is necessary, possibly resulting in a faster and more efficient system compared to using FP for every pose estimate.

\section{Conclusion}
In this work, we investigated a structured alternative to general pose estimation methods for robotic manipulation. Our approach decomposes pose estimation into four stages: initial pose estimation, failure detection, error source attribution, and targeted mitigation strategies. We demonstrated that such a design can significantly elevate grasp success, achieving comparable success rates to FoundationPose \cite{bowen_foundationpose_2024} as a state-of-the-art general pose estimation method while relying only on ICP as a simple and lightweight pose estimator. 

In doing so, we significantly improve real-world error attribution accuracy compared to previous work \cite{gaus2025human} by deploying PointBERT \cite{yu2021pointbert} as an error attribution model. Additionally, we provide a lightweight grasp success prediction MLP and a custom point cloud reconstruction module.

The results show that robust pose estimation for real-world robotic manipulation tasks does not necessarily require computationally expensive and slow foundational general pose estimation models. Instead, explicitly identifying and mitigating error sources can significantly improve the robustness of substantially simpler and more lightweight pose estimators.

Future work should apply this approach to state-of-the-art pose estimators to build uncertainty-aware frameworks, improving current performance levels. Furthermore, this work motivates a deeper investigation of error sources in current pose estimation methods with the goal to detect and classify error sources. Based on this, targeted mitigation strategies should be developed and integrated into adaptive recovery frameworks.


\bibliographystyle{IEEEtran}  
\bibliography{bib/bibliography}      

\end{document}